# Review of feedback in Automated Essay Scoring

You-Jin Jong[1], Yong-Jin Kim[2], Ok-Chol Ri[1]

[1] Kum Sung Middle School Number 2, Pyongyang, 999093, D.P.R of Korea
[2] Faculty of Mathematics, KIM IL SUNG University, Pyongyang, 999093, D.P.R of Korea

Correspondence should be addressed to Yong-Jin Kim: kyj0916@126.com

## Abstract

The first automated essay scoring system was developed 50 years ago. Automated essay scoring systems are developing into systems with richer functions than the previous simple scoring systems. Its purpose is not only to score essays but also as a learning tool to improve the writing skill of users. Feedback is the most important aspect of making an automated essay scoring system useful in real life. The importance of feedback was already emphasized in the first AES system. This paper reviews research on feedback including different feedback types and essay traits on automated essay scoring. We also reviewed the latest case studies of the automated essay scoring system that provides feedback.

## 1. Introduction

Due to the current COVID-19 crisis, online education systems are becoming more active and more important. In addition, with the development of science and technology, changes are taking place in the way people learn. There are so many learning materials that make self-study one of the useful learning methods.

Learning foreign languages has become commonplace. Learning a foreign language is not only to satisfy our interests. Speaking a second or third language is an essential skill in today's multicultural society. Writing is an important part of language learning. Writing skills assessments are included in all language tests.

Automated Essay Scoring (AES) is the task of assessing writing skills and scoring essays without human intervention. The process of manually scoring essays is complex and time-consuming. Even with a fixed scoring guide, the scoring process is affected by individual factors such as mood and personality, and the assigned score is subjective and unreliable. In addition, students cannot self-study if others have to manually score their writing. Automatic scoring of essays has been proposed as a solution to manual scoring. It can be used in language tests and self-study.

Several reviews on AES systems have been published and mentioned in detail (Hussein, Hassan, & Nassef, 2019; Ke, & Ng, 2019; Klebanov, & Madnani, 2020; Lagakis, & Demetriadis, 2021; Ramesh, & Sanampudi, 2022; Uto, 2021). In Uto 2021, AES systems using deep neural networks were reviewed. In Ramnarain-Seetohul, Bassoo, and Rosunally 2022, more specifically, the similarity measurement methods used to implement AES systems were reviewed.

In Uto 2021, they reviewed the neural network models for AES by classifying them into four types: prompt-specific holistic scoring, prompt-specific trait scoring, cross-prompt holistic scoring, and cross-prompt trait scoring. This classification is very interesting, and the fourth type (cross-prompt trait scoring) can be seen as the direction of the current research. Cross-prompt trait scoring means predicting multiple trait-specific scores for each essay in a cross-prompt setting in which essays written for non-target prompts are used to train an AES model. It is a system that can cope with small data and provide feedback to users.

Feedback gives users an interpretation of the score, so that they are convinced about the score, and can also provide an answer on what to do to improve the score. In the current state of research, a system that does not provide feedback can no longer be put to practical use, and those who have used a feedback system no longer want to use a system without feedback. Therefore, considering the importance of feedback in the existing AES system, we reviewed only feedback in the AES task. Feedback was discussed as part of previous reviews. The feedback was reviewed in Lagakis, and Demetriadis 2022. They reviewed the recently published scientific literature on AES systems that provide feedback and their impact on learning, and the user's attitude towards using such systems in the learning procedure. In this paper, we reviewed the feedback in a brand way and analyzed the latest case studies.

We searched Google Scholar using the keyword "Automated Essay Scoring Feedback". Even if a word in the keyword is not included, all papers related to AES are extracted because they are sorted according to their relevance. In addition, all papers cited in the feedback part of previous reviews were also collected (Hussein et al, 2019; Ke et al, 2019; Klebanov et al, 2020; Lagakis et al, 2021; Lagakis et al, 2022; Ramesh et al, 2022; Ramnarain-Seetohul et al, 2022; Uto, 2021).

The remainder of this paper is organized as follows. In the Feedback Types Section, we reviewed various feedback types used in AES systems. In the Datasets Section and Essay Traits Section, we reviewed datasets and essay traits used for the feedback study. In the Implementations Section, implementations of feedback systems are reviewed and compared to systems without feedback. In the Case Studies Section, recently conducted case studies were reviewed. Finally, we conclude by discussing the problems and limitations of the feedback study.

## 2. Feedback Types

Many AES systems that provide feedback have been studied and are widely used in real life. Feedback on essays can be viewed as an evaluation of the traits of the essay rather than the entire essay. Upon reviewing feedback systems, we concluded that assigning scores to traits is the basic type of feedback. In other words, this type of feedback can be used to provide other types of feedback.

We reviewed various feedback types provided by existing AES systems. Most AES systems provide formative feedback. What makes these systems intelligent is their ability to return valid, formative feedback that students can apply to improve their writing skills.

In Shehab, Elhoseny, and Hassanien 2016, the system provided feedback for four traits in the form of a declarative sentence. For example, "Your essay might not be relevant to the assigned topic" or "Your essay shows less development of a theme than other essays written on this topic". In Criterion (Burstein, Chodorow, & Leacock, 2004), feedback was provided not only for low-score essays but also for high-score essays. "Well organized, with strong transitions helping to link words and ideas." and "Varies sentence structures and makes good word choices; very few errors in spelling, grammar, or punctuation." are the feedback sentences for an essay assigned a high score. "Lacks organization, and is confused and difficult to follow; may be too brief to assess the organization." and "Little or no control over sentences, and incorrect word choices may cause confusion; many errors in spelling, grammar, and punctuation severely hinder reader understanding." are the feedback sentences for an essay assigned a low score. This feedback can be provided by matching various declarative sentences with different trait scores.

In Liu, Li, Xu, and Liu 2016, the system provided feedback for seven traits in the form of an interrogative sentence. For example, "Are the ideas used in the essay relevant to the question?" or "Are the ideas developed correctly?". Like the above, this can also be provided by matching various interrogative sentences with low trait scores.

In Wiring pal (Roscoe, & McNamara, 2013), the feedback was provided more comprehensively. The feedback provided in Wiring pal is also in the form of a declarative or interrogative sentence, and if it is different from the feedback mentioned above, it tells in more detail how to improve the trait scores. For example, "Could a stranger understand your ideas without further explanation?", "Remember that you can always add more elaboration to make sure that the reader understands your point of view!" and "Don't forget to start each body paragraph with a short and simple topic sentence".

In Woods, Adamson, Miel, and Mayfield 2017, they picked good and bad sentences for each trait, in addition to giving them a score of 1 to 4 for four traits. This feedback was also provided by calculating the difference in scores between cases with and without the certain sentence.

As can be seen, various types of feedback can be provided through trait scoring. Therefore, assigning scores to traits is the most basic, effective, and important type of feedback.

## 3. Datasets

This section describes the datasets used in the feedback study. The purpose of this section is to introduce the various traits used as feedback through a description of the datasets. These traits are described in detail in the Essay Traits Section. Most researchers conducted and evaluated experiments on their datasets (Crossley, Kyle, K., & McNamara, 2015; Liu et al, 2016; Ng, Bong, Lee, & Sam, 2015; Ng, Bong, Sam, & Lee, 2019; Shehab et al, 2016). There are a few public datasets with feedback annotations. Public datasets are very important and useful. First, new researchers in the field do not have to worry about creating a new dataset. Next, researchers can perform a fair comparison of different methods and models by using those datasets.

Unlike previous reviews, we reviewed datasets in detail. We focused on the feedback provided by the datasets. We reviewed only datasets that have trait scores. From the conclusion in the Feedback Types Section, these datasets are very useful in the feedback study. For example, the Cambridge Learner Corpus-First Certificate in English exam (CLC-FCE) (Yannakoudakis, Briscoe, & Medlock, 2011) provides for each essay both its holistic score and the manually tagged linguistic error types using a taxonomy of approximately 80 error types. We did not describe this dataset, which has been mentioned in previous reviews.

## 3.1 ASAP/ASAP++

In 2012, Kaggle hosted the Automated Student Assessment Prize (ASAP) competition to assess the capabilities of AES systems. The ASAP dataset is built with essays written by students ranging in grade levels from grade 7 to grade 10. There are approximately 13,000 essays corresponding to 8 prompts. The specific dataset information is presented in Table 1. Each prompt has a different score range and the number of essays.

Table 1: Statistics of ASAP dataset.

| Prompt | Number of Essays | Score Range |
|--------|------------------|-------------|
| 1 | 1783 | 2-12 |
| 2 | 1800 | 1-6 |
| 3 | 1726 | 0-3 |
| 4 | 1772 | 0-3 |
| 5 | 1805 | 0-4 |
| 6 | 1800 | 0-4 |
| 7 | 1569 | 0-30 |
| 8 | 723 | 0-60 |

In ASAP, 6 of the 8 prompts only have overall scores. Only 2 (prompts 7 and 8) of them have scores for individual essay traits such as content, organization, style, and so on. In 2018, Mathias and Bhattacharyya presented a manually annotated dataset ASAP++ for the other six prompts (Mathias, & Bhattacharyya, 2018).

Since the trait scores were rated by different people, there is a slight difference between the trait scores of prompts 7 and 8 and the trait scores of prompts 1-6. For example, the overall score of prompts 7 and 8 is calculated from the trait scores, and others are not. The trait scores of the other six prompts are independent of the overall score. ASAP has been widely used in overall scoring. More than 90% of previous studies were evaluated using ASAP (Ramesh, & Sanampudi, 2022). ASAP++ is also frequently used in trait scoring (Mathias et al., 2018; Ridley, He, Dai, Huang, & Chen, 2021). For this reason, this dataset is explained first. Table 2 shows the traits of each prompt provided in ASAP/ASAP++. For the type of essay, "A", "SD", and "N" mean "Argumentative", "Source-dependent", and "Narrative", respectively.

Table 2: Essay traits provided in ASAP/ASAP++.

| Prompt | Essay Type | List of Essay Traits | Score Range |
|--------|-----------|----------------------|-------------|
| 1 | A | Content, Organization, Word Choice, Sentence Fluency, Conventions | 1-6 |
| 2 | A | Content, Organization, Word Choice, Sentence Fluency, Conventions | 1-6 |
| 3 | SD | Content, Prompt Adherence, Language, Narrativity | 0-3 |
| 4 | SD | Content, Prompt Adherence, Language, Narrativity | 0-3 |
| 5 | SD | Content, Prompt Adherence, Language, Narrativity | 0-4 |
| 6 | SD | Content, Prompt Adherence, Language, Narrativity | 0-4 |
| 7 | N | Content, Organization, Style, Conventions | 0-3 |
| 8 | N | Content, Organization, Voice, Word Choice, Sentence Fluency, Conventions | 1-6 |

An argumentative(persuasive) essay is a genre of writing that requires the student to investigate a topic, collect, generate, and evaluate evidence, and establish a position on the topic concisely. A narrative essay is a creative genre of writing in which students have to tell a story. A source-dependent essay is a genre of writing that students read an article before writing and write essays referencing this article.

## 3.2 ICLE

The International Corpus of Learner English (ICLE) is a dataset of essays written by upper-intermediate and advanced learners. Founded and coordinated by Sylviane Granger of the University of Louvain, the dataset is the result of almost 30 years of international collaborative activity between a large number of universities.

When it first came out, this dataset did not have trait scores. Since 2010, some researchers have annotated some essays in the dataset. These traits include Organization (Persing, Davis, & Ng, 2010), Thesis Clarity (Persing, & Ng, 2013), Prompt Adherence (Persing, & Ng, 2014), Argument Strength (Persing, & Ng, 2015), and Thesis Strength (Ke, Inamdar, Lin, & Ng, 2019).

The ICLE dataset continues to grow, but essays having trait scores do not. Table 3 shows the number of essays annotated for traits so far.

Table 3: ICLE dataset traits and number of essays.

| No | Essay Type | Trait | Number of essays | Score Range |
|----|-----------|-------|------------------|-------------|
| 1 | A | Organization | 1003 | 1-4(0.5) |
| 2 | A | Thesis Clarity | 830 | 1-4(0.5) |
| 3 | A | Prompt Adherence | 830 | 1-4(0.5) |
| 4 | A | Argument Strength | 1000 | 1-4(0.5) |
| 6 | A | Thesis Strength | 1021 | 1-6 |

In Table 3, the score intervals are given in parentheses. For example, 1-4 (0.5) indicates that there are 7 scores from 1 to 4 at 0.5 intervals. For the last trait, they defined ten sub-traits that could be used as feedback to that trait and assigned them a score from 1 to 3. These ten sub-traits are Arguability, Specificity, Clarity, Relevance to Prompt, Conciseness, Eloquence, Confidence, Direction of Development, Justification of Opinion, and Justification of Importance/Interest.

## 3.3 AAE

Argument Annotated Essays (AAE) contains 402 essays taken from essayforum2, a site that offers feedback to students who wish to improve their ability to write argumentative essays (Stab, & Gurevych, 2014). Each essay was annotated with its argumentative structure (for example, argument components such as claims and premises as well as the relationships between them). In 2018, Carlile et al. annotated some essays with Argument Persuasiveness scores (Carlile, Gurrapadi, Ke, & Ng, 2018). They randomly selected 102 essays from the dataset and scored their persuasiveness.

They also defined five sub-traits that could be used as feedback for that trait. These five sub-traits are Eloquence, Evidence, Claim and MajorClaim Specificity, Premise Specificity, and Relevance.

## 3.4 RTA

Correnti et al. created a Response-to-Text Assessment (RTA) dataset used to check student writing skills in four directions (Analysis, Organization, Style, Evidence, and MUGS) (Correnti, Matsumura, Hamilton, & Wang, 2013). The 4-8 grade students gave their responses to the RTA. MUGS refers to mechanics, usage, grammar, and spelling, they use the trait MUGS to reflect the writer's ability to adhere to grade-appropriate standard writing conventions.

Table 4 provides comprehensive information about the datasets.

Table 4: Datasets with feedback annotations.

| No | Dataset | Type | Trait | Number of essays | Score Range |
|---|---|---|---|---|---|
| 1 | ASAP/ ASAP++ | A | Content Organization Word Choice Sentence Fluency Convention | 3583 | 1-6 |
| | | SD | Content Prompt Adherence Language Narrativity | 7103 | 0-3 or 0-4 |
| | | N | Content Organization Conventions Style Voice Word Choice Sentence Fluency | First 3: 2292 Style: 1569 Last 3: 723 | 0-3 or 1-6 |
| 2 | ICLE | A | Organization | 1003 | 1-4(0.5) |
| | | | Thesis Clarity | 830 | 1-4(0.5) |
| | | | Prompt Adherence | 830 | 1-4(0.5) |
| | | | Argument Strength | 1000 | 1-4(0.5) |
| | | | Thesis Strength | 1021 | 1-6 |
| 3 | AAE | A | Argument Persuasiveness | 102 | 1-6 |
| 4 | RTA | SD | Analysis Organization Style Evidence MUGS | 5046 | 1-4 |

## 4. Essay Traits

Essay traits can be divided into two types: content-dependent(high-level) traits and content-independent(low-level) traits. For example, traits such as grammar, and word spelling are traits that are not related to the content of the essay and do not help improve writing skills. Content-dependent traits are more difficult to evaluate than content-independent traits. Content-dependent traits are an important part of a human assessor's evaluation and are still very difficult to estimate effectively using AES systems.

From the previous section, we have seen many traits in the four datasets. Some traits in different datasets are similar, and some traits can be viewed as combinations of other traits. There are also instances where a trait with the same name is used for different meanings in different datasets. For example, the trait Style has different meanings in ASAP and RTA. For some traits, if it is difficult for users to understand the meaning of that trait, it cannot be a good trait. Therefore, among the various traits, those that are very important and can help improve writing skills were reviewed.

It should be noted that the traits listed below are not available for all types of essays. For example, traits such as argument strength and thesis clarity are suitable for argumentative essays but not for source-dependent essays.

## 4.1 Content-dependent traits

*Organization* refers to the structure of the essay. An essay with a high organization score is structured in a way that logically develops an argument (Correnti et al., 2013; Hussein, Hassan, & Nassef, 2020; Mim, Inoue, Reisert, Ouchi, & Inui, 2019; Mathias et al., 2018; Mathias et al., 2020; Persing et al., 2010; Ridley et al., 2021).

*Coherence* refers to the semantic relatedness among sentences and the logical order of concepts and meanings in a text.

*Cohesion* refers to the use of linguistic devices (indicators, coreference, substitution, ellipsis, and so on) that hold a text together.

*Narrativity* refers to the coherence and cohesion of the response to the prompt (Mathias et al., 2018; Mathias et al., 2020; Ridley et al., 2021).

*Prompt adherence* refers to the relatedness between the response and the prompt. An essay with a high prompt adherence score remains on the topic introduced by the prompt and is free of irrelevant digressions (Louis, & Higgins, 2010; Mathias et al., 2018; Mathias et al., 2020; Persing et al., 2014; Ridley et al., 2021).

*Thesis clarity* refers to the ability to explain the thesis of an essay clearly. An essay with a high thesis clarity score presents its thesis in a way that is easy for the reader to understand (Persing et al., 2014).

*Argument strength (Argument persuasiveness)* refers to the strength of the argument an essay makes for its thesis. An essay with a high argument strength score presents a strong argument for its thesis and would convince most readers (Carlile et al., 2018; Ke, Carlile, Gurrapadi, & Ng, 2018; Mim et al., 2019; Persing et al., 2015).

*Thesis Strength* refers to the strength of the thesis statement in an argumentative essay. A strong thesis statement can help lay a strong foundation for the rest of the essay by organizing its content, improving its comprehensibility, and ensuring its relevance to the prompt. In contrast, an essay with a weak thesis statement lacks focus (Ke et al, 2019).

*Evidence* refers to the degree to which writers select and use details, including direct quotations from the text to support their key idea (Correnti et al., 2013).

*Voice* refers to the ability to choose an appropriate voice for the topic, purpose, and audience. An essay with a high voice score demonstrates a deep commitment to the topic, and there is an exceptional sense of "writing to be read" (Mathias et al., 2020).

*Analysis* refers to the ability to demonstrate a clear understanding of the purpose of the literary work and to make valid and perceptive conclusions that inform an insightful response to the prompt (Correnti et al., 2013).

*Sentence Fluency* refers to the flow and rhythm of the essay. An essay with a high sentence fluency score is characterized by a natural, fluent sound and sentence structure that enhances meaning (Mathias et al., 2018; Mathias et al., 2020; Ridley et al., 2021).

### 4.2 Content-independent traits

*Word Choice* refers to the usage of words that convey the intended message in an interesting, precise, and natural way appropriate to the audience and purpose. An essay with a high word choice score is characterized by natural and not overdone words and vocabularies that evoke clear images (Mathias et al., 2018; Mathias et al., 2020; Ridley et al., 2021).

*Language* refers to the quality of the grammar and spelling in the response (Mathias et al., 2018; Mathias et al., 2020; Ridley et al., 2021).

*Conventions/Mechanics/Usage/Grammar/Spelling (MUGS)* refers to the control of standard writing conventions for grammar, usage, spelling, capitalization, punctuation, and so on. An essay with high conventions score is characterized by effective use of punctuation, correct spelling, correct grammar, and usage that contribute to clarity and style (Correnti et al., 2013; Hussein et al., 2020; Mathias et al., 2018; Mathias et al., 2020; Ridley et al., 2021).

## 5. Implementations

### 5.1 Studies in Model

In the Feedback Types Section, we have already analyzed the importance of trait scoring in feedback. The AES system that provides feedback can be viewed as assigning multiple scores (trait scores) instead of assigning one overall score. Therefore, the implementation methods are divided into a method based on feature engineering and a method using a deep neural network, similar to the implementation method of the overall score AES system. The difference is that since we care about multiple scores, we are more concerned with the relationship between those scores.

The datasets used in AES came out a long time ago, but the model using the deep neural network has been used since 2016 (Taghipour, & Ng, 2016; Alikaniotis, Yannakoudakis, & Rei, 2016) and has become the current research direction.

**Use the same model to get trait scores**

In Hussein et al. 2020, they trained the model for prompt 7 of the ASAP, using the model from Taghipour et al. 2016. Figure 1 shows the model from Taghipour et al. 2016. Many models were modified from this model and their performance was also compared with this model. The following models to obtain trait scores are also based on this model, we explained this model in detail. This is the simplest and most representative model. It generates a representation of the input essay and obtains value from it.

The lookup table layer projects each word in an essay into dimensional space. An essay $W$ is represented by every word's one-hot vector $w_i (i = \overline{1..N})$, and the output of the lookup table layer is calculated by Equation 1:

$$LT(W) = (Ew_1, Ew_2, ..., Ew_N) \tag{1}$$

where $E$ is the word embedding matrix and will be learned during training.

The convolution layer extracts local features from the essay and the recurrent layer generates a representation for an essay. In the mean over time layer, the sum of outputs of the recurrent layer is divided into the essay length. Let $H = (h_1, h_2, ..., h_N)$ be the outputs of the recurrent layer. The function of the mean over time layer is defined by Equation 2:

$$MoT(H) = \frac{1}{N} \sum_{t=1}^{N} h_t \tag{2}$$

In the last linear layer, they used the sigmoid function to get a score in the range of (0,1). The sigmoid function is given by Equation 3:

$$s(H) = sigmoid(w \cdot MoT(H) + b) \tag{3}$$

Therefore, they normalized all scores to [0,1] before training the model.

From this original model, trait scores were obtained by adding several outputs to the last output layer in addition to the overall score (see Figure 2). The trait scores were calculated in the same way as the overall score. They attempted to add a dense layer to the original model, but the performance did not improve.

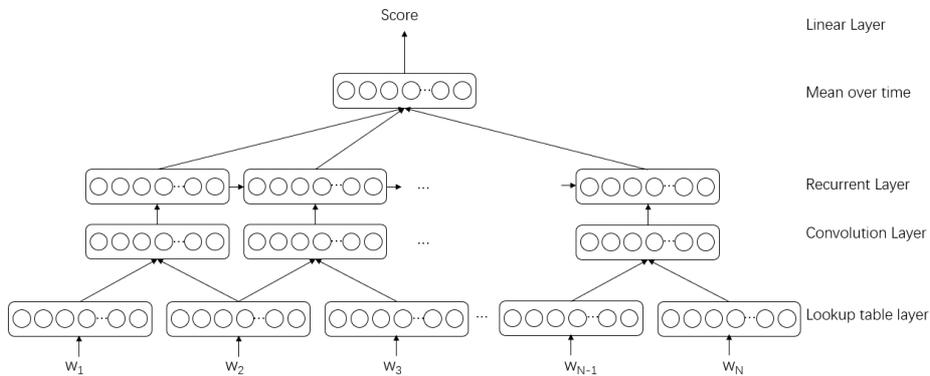

Figure 1: Architecture of the model in Taghipour et al. 2016

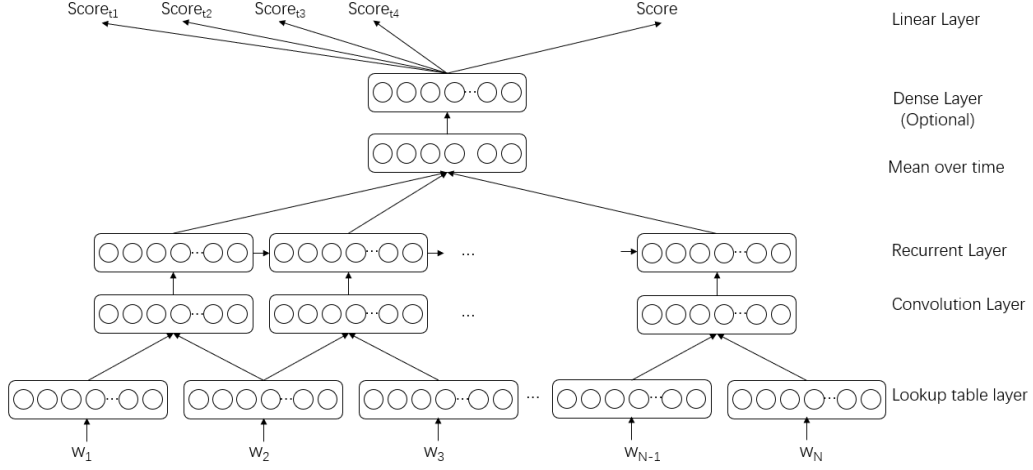

Figure 2: Architecture of the model in Hussein et al. 2020

In Mathias et al. 2020, trait scores were obtained using the model from Dong, Zhang, and Yang 2017. In Dong et al. 2017, they achieved good performance in the overall ASAP scoring using CNN and the attention layer to generate representation vectors for each word and sentence (see Figure 3). There are two attention layers in this model and they work in the same way. For example, the last attention layer to get the final essay representation is calculated as follows:

$$a_i = \tanh(\boldsymbol{W}_a \boldsymbol{h}_i + \boldsymbol{b}_a) \tag{4}$$

$$\alpha_i = \frac{e^{\boldsymbol{w}_\alpha a_i}}{\sum e^{\boldsymbol{w}_\alpha a_j}} \tag{5}$$

$$\boldsymbol{o} = \sum \alpha_i \boldsymbol{h}_i \tag{6}$$

where $\boldsymbol{W}_a$, $\boldsymbol{w}_\alpha$ are weight matrix and vector respectively, $\boldsymbol{b}_a$ is the bias vector, $a_i$ is attention vector for a sentence, and $\alpha_i$ is the attention weight of the sentence. $\boldsymbol{o}$ is the final essay representation, which is the weighted sum of all the sentence vectors.

The difference from the above method is that several scores are trained independently rather than in a single model. The advantage is that the performance can be improved by training a single model for each trait score, and the disadvantage is that the training time is long and the relationship between the feature scores cannot be considered.

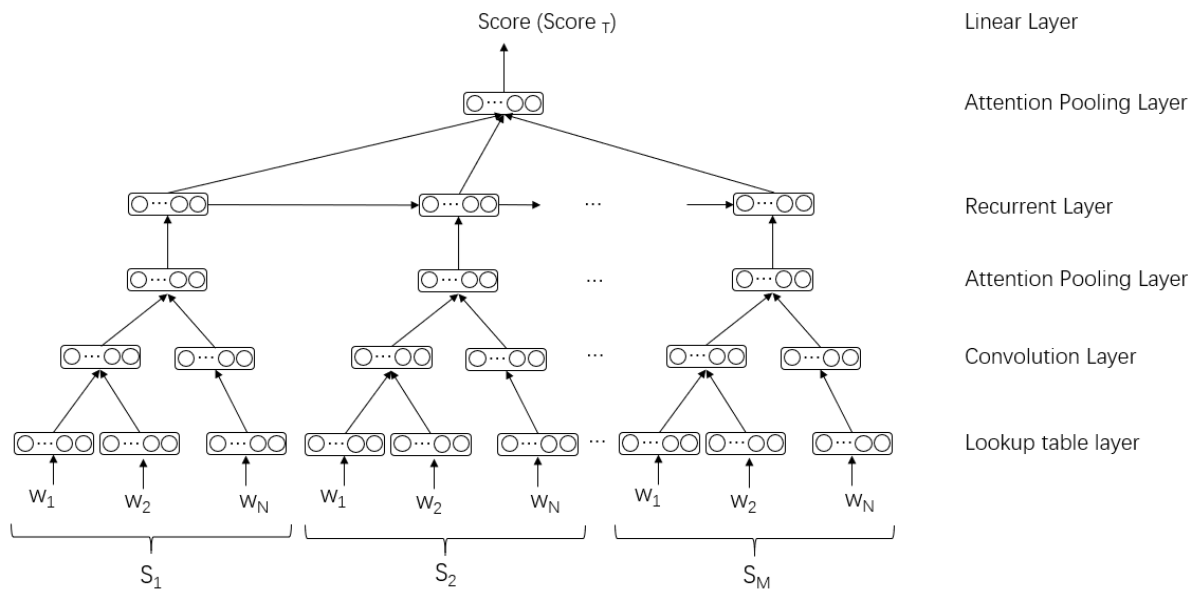

Figure 3: Architecture of the model in Dong et al. 2017

In Lu, and Cutumisu 2021, an experiment was also evaluated with ASAP, but the feedback was generated from the overall score, not from the trait scores. In ASAP, there are rubric guidelines for overall scores. For example, for score 1, the description is "An undeveloped response that may take a position but offers no more than very minimal support. Typical elements: Contains few or vague details, Is awkward and fragmented, May be difficult to read and understand, May show no awareness of audience." And for score 6, the description is "A well-developed response that takes a clear and thoughtful position and provides persuasive support. Typical elements: Has fully elaborated reasons with specific details, Exhibits strong organization, Is fluent and uses sophisticated transitional language, May show a heightened awareness of audience."

They designed and trained a model to obtain the overall scores using various neural networks. The final feedback is obtained by using the feedback generator learned through the overall score and rubric guidelines.

This implementation is different from the above implementations in that it does not depend on the trait score, however, this method can also be used for the trait score. From the above implementations, we know that trait scores can be obtained using the same model. After obtaining the trait scores, if you have the corresponding rubric guidelines, you can get more detailed feedback. In the Feedback Type Section, we explained some feedback types that give the same feedback sentence for the same trait score. By generating feedback differently, it is possible to ensure freshness and eliminate boredom while using the system.

**Use the modified model to get trait scores**

In Ridley et al. 2021, they modified the model from Ridley, He, Dai, Huang, & Chen 2020 (see Figure 4). These models are used for the cross-prompt AES task. Directly using word embeddings is not suitable for the cross-prompt task, therefore, they get part-of-speech (POS) embeddings from the essay. They also use non-prompt-specific features such as Readability and Text Complexity to represent the quality of an essay from different dimensions.

In Ridley et al. 2021, they modified the model such that all traits share the lower part of the model. But at the upper part, they are calculated independently (see Figure 5). After calculating independently, the attention layer was added by considering the relationship between the traits. This can be considered a relatively reasonable implementation by combining the two above implementations.

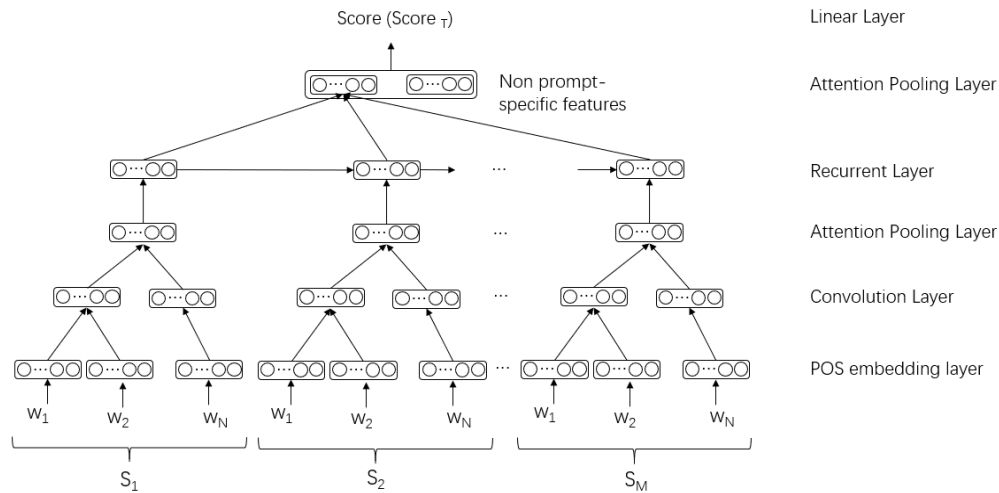

Figure 4: Architecture of the model in Ridley et al. 2020 (it was also modified based on the model from Dong et al. 2017)

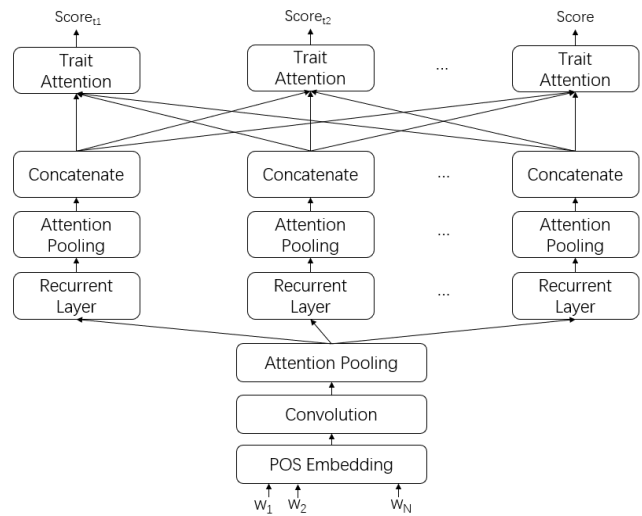

Figure 5: Architecture of the model in Ridley et al. 2021

## 5.2 Studies in System

Since the feedback system is not a simple overall scoring system, many aspects need to be considered and improved compared to the overall scoring system in actual implementation. For this reason, research on the scoring system, which has only a few traits described in the Essay Traits Section, is constantly in progress.

In Ke et al. 2018, the system allows teachers to collect samples of scored essays to be trained to score newly entered essays. Teachers can set tasks, keep track of student progress, provide additional feedback, and rectify the generated scores. Students can practice writing essays and demand feedback at any point of their essay writing process so that the system can provide scores by paragraph and the entire essay.

In Tashu, and Horváth 2019 and Ye, and Manoharan 2019, they proposed and implemented AES systems that provide personalized feedback. These systems are designed in a similar way. The teacher provides feedback on some essays, and the feedback is used to provide feedback on other similar essays as well. This method determines the correctness of students' answers according to the meaning of the answers instead of comparing the words in the answers with those in the correct answer.

In Tashu, et al. 2019, the system will allow teachers to interact with the system, give feedback on the student's essays in the form of textual comments, and provide recommendations to other similar essays based on those essays. To compute the semantic similarity and to provide feedback recommendations, they used neural word embedding and relaxed word mover's similarity.

In Ye, et al. 2019, the system partitions the students' essays into groups according to their semantic meanings. Teachers provide feedback on a few essays in each group, which covers the typical issues demonstrated by the essays in the group. The feedback is propagated to the other essays in the same group. To help teachers to provide personalized feedback to students promptly, the $k$-mean algorithm is applied to the vectors to partition the students' essays into groups according to their meanings.

## 6. Case Studies

This section describes several recent case studies. The AES systems used in these studies, the educational level of the participating users, and the number of participants varied. In addition, the native language of some users is not English which is a language supported by the AES system. Although the experiment was conducted in these different environments, the users' final attitudes toward the system were mostly positive and some were neutral.

What is surprising is that in this field, case studies have been conducted more than actual performance improvement studies. Since the purpose of feedback is to improve students' writing skills, it is very important to determine whether this objective has been accurately achieved. From this point of view, in actual use, it is appropriate for others to evaluate the usefulness of systems designed and advertised by developers.

Table 5 shows all the case studies we analyzed.

Table 5: Case studies of AES systems that provide feedback.

| | System | Time | Participants | Native | Education | Content |
|---|---|---|---|---|---|---|
| Wilson, and Roscoe 2020 | PEG Writing, Google Docs | 7 months | 114 students 3 teachers | Yes | High school | Studied the effectiveness of the system using writing self-efficacy, holistic writing quality, performance on a test, and |
| | | | | | | teachers' perceptions of the social validity. |
| Lee (2020) | Criterion | 1 year | 2 students | No | University | Studied how university students improved their writing skills using a mixed-methods research design to track. |
| Link, Mehrzad, and Rahimi 2022 | Criterion | 1 semester | 12 students 16 teachers | No | University | Studied whether the system can assist teachers by allowing them to devote more feedback to high-level writing skills by comparing two writing classes assigned to either a system and teacher feedback condition or a teacher-only feedback condition. |
| Jiang, and Yu 2022 | Pigai | 16 weeks | 9 students | No | University | Studied how students employ resources and strategies to use feedback using activity theory and the appropriation construct. |
| Huang, and Renandya 2020 | Pigai | 16 weeks | 67 students | No | University | Studied the system's impact on learners at a lower language level, and their |

| | | | | | | perceptions of the system. |
|---|---|---|---|---|---|---|
| Jiang, Yu, and Wang 2020 | Pigai | 1 semester | 11 teachers | No | University | Studied the system's impact on teachers' feedback. |
| Li 2021 | Criterion | 1 semester | 3 teachers 80 students | No | University | Studied the roles in a system-aided environment and their potential impact on writing development. |
| Wilson, Ahrendt, Fudge, Raiche, Beard, and MacArthur 2021 | MI Write | 1 semester | 90 teachers 3000 students | Yes | Elementary school | Studied the attitudes towards the system, and system use experiences using a focus group methodology. |
| Wilson, Huang, Palermo, Beard, & MacArthur 2021 | MI Write | 1 year | 1935 students 135 teachers | Yes | Elementary school | Studied the extent to which aspects of the system were implemented, teacher and student attitudes towards the system, and associations between system usage and students' outcomes. |
| Huang, and Wilson 2021 | MI Write | 1 year | 431 students 44 teachers | Yes | Elementary school | Studied growth in writing quality associated with feedback and whether prolonged usage of the system was associated |

| | | | | | | with gains in students' independent performance. |
|---|---|---|---|---|---|---|
| Razak, Lotfie, and Zamin 2021 | WriteLab | 1 semester | 80 students | No | University | Studied the perceptions of learners on the use of the system to improve their writing skills using a quasi-experimental research design. |
| Zhai, and Ma 2022 | Pigai | 1 semester | 448 students | No | University | Studied an extended technology acceptance model to identify the environmental, individual, educational, and systemic factors that influence students' acceptance of feedback and examined how they affect students' usage intention. |
| Gao 2021 | Pigai | - | 104 students' essays | No | University | Studied the feedback quality of the system and students' perceptions of the feedback use. |
| Ranalli 2021 | Grammarly | - | 6 students | No | University | Studied the extent to which engagement supports learning from, and is influenced by |

| | | | | | | trust in, system feedback. |
| McCarthy, Roscoe, Allen, Liken, and McNamara 2022 | Writing Pal | 4 sessions | 119 students | Both | High school | Studied the extent to which adding spelling and grammar checkers support writing and revision compared to providing writing strategy feedback alone. |

In Gao 2021 and Ranalli 2021, because they researched already existing essays, it did not take a long time to use the systems. In McCarthy et al. 2022, they did not conduct a long-term case study, it took place over 4 sessions. There is also the case of conducting a series of different case studies for the same system (Huang et al., 2021; Wilson et al., 2021; Wilson et al., 2021). Usage was higher for non-native speakers. Experiments were conducted in elementary and high schools for native speakers, and universities for non-native speakers. The AES system is increasingly used for learning among students whose native language is not English. In the case of research on the improvement of writing skills, the minimum research time was half a year.

## 7. Conclusions

First, based on the analysis of feedback systems and previous studies, we concluded that assigning scores to traits is the most basic and important feedback type. Basic means that this type of feedback can be used to provide other types of feedback. Second, we believe that the traits necessary to evaluate writing have been sufficiently studied and analyzed. In recent years, there have been no more traits that are stand-alone (essentially different from traits that have already been used) and that can be considered useful. Third, many AES systems that provide feedback have been used, and the attitudes of users toward these systems are very positive. Users of these systems include all kinds of students and teachers from elementary school to university and a wide range of people.

Datasets are very important, but in the AES task, datasets with large data and many traits are lacking. Furthermore, many datasets are not available to the public because they are widely used for commercial purposes. Another problem is that, in many studies, final scores are not obtained from the trait scores, but are obtained independently using the same type of method. This is an important issue that should be addressed in the future. If the relationship between a trait's score and the overall score is not clear, it may also result in a decrease in the overall score of the essay modified to increase the trait score. From this perspective, the generation of a dataset requires ensuring consistency from the overall score to the trait score.

Many case studies have been conducted and the effectiveness and validity of feedback systems have been verified. However, because there are no standard methods and criteria to evaluate the quality of feedback or the convenience of the system, many case studies have been conducted on the same system and their conclusions have been drawn.

Even if it is a basic feedback type to give a score to a trait, AES systems can provide more diverse and rich feedback based on this. Although overall scoring and trait scoring are similar from the point of scoring, there is still room for improvement in feedback. Diversity in feedback is a direction for future research. As mentioned in the Implementations Section, the diversity here refers to diversity in formative feedback itself and diversity in terms of the realization of how to effectively apply this feedback to the system. Even with the same feedback, the system can be designed and implemented in a convenient way.

Therefore, the AES system will be used more widely if the performance of evaluating the traits is improved and the feedback types are enriched in the future.